\documentclass{article} 
\usepackage{iclr2018_conference,times}
\usepackage{hyperref}
\usepackage{url}
\usepackage[utf8x]{inputenc}
\usepackage{graphicx}     
\usepackage{listings}
\usepackage{makecell}

\title{Generating Wikipedia by Summarizing Long Sequences}

\author{Peter J. Liu\thanks{Joint first-authors. Ordered randomly.} , Mohammad Saleh\footnotemark[1], \\
\textbf{Etienne Pot\thanks{Work done as a member of the Google Brain Residency (g.co/brainresidency)} , Ben Goodrich, Ryan Sepassi, \L{}ukasz Kaiser, Noam Shazeer} \\
Google Brain\\
Mountain View, CA\\
\texttt{\{peterjliu,msaleh,epot,bgoodrich,rsepassi,lukaszkaiser,noam\}@google.com} \\
}

\newcommand{\coderepo}{\url{https://goo.gl/wSuuS9}}

\iclrfinalcopy 

\begin{document}

\DeclareRobustCommand{\model-wiki-input-cap}[4]{
\begin{figure}[h]
\fbox{%
    \parbox{\textwidth}{%
        \textbf{Model:}
        #1
        \par
        \textbf{Wikipedia:}
        #2
        \par
        \textbf{Input:}
        #3
}%
}%
\caption{#4}
\end{figure}
}

\maketitle

\begin{abstract}
    We show that generating English Wikipedia articles can be approached as a multi-document
    summarization of source documents.
    We use extractive summarization to coarsely identify salient information
    and a neural abstractive model to
    generate the article.
    For the abstractive model, we introduce a decoder-only architecture that can
	scalably attend to very long sequences,
	much longer than typical encoder-decoder architectures
	used in sequence transduction.
    We show that this model can generate fluent, coherent multi-sentence
        paragraphs and even whole Wikipedia articles.
    When given reference documents, we show it can extract relevant
        factual information
        as reflected in perplexity, ROUGE scores and human evaluations.
\end{abstract}

\section{Introduction}
The sequence-to-sequence framework has demonstrated success in
natural-language sequence
transduction tasks such as machine translation.
More recently,
neural techniques have been applied to do single-document,
abstractive (paraphrasing)
text summarization of news
articles (\cite{Rush2015}, \cite{nallapati2016abstractive}). In this prior work, the input to
supervised models ranged from the first sentence to the entire
text of an article, and they are trained end-to-end to predict reference summaries.
Doing this end-to-end requires a significant number of parallel
article-summary pairs since language understanding
is a pre-requisite to generate fluent summaries.

In contrast, we consider the task of multi-document summarization, where
the input is a collection of related documents from which a summary is
distilled. Prior work has focused on extractive summarization,
which select sentences or phrases from the input to form the summaries, rather
than generating new text.
There has been limited application of abstractive neural methods and one
possible reason is the paucity of large, labeled datasets.

In this work,
we consider English Wikipedia as a supervised machine learning task
for multi-document summarization where the input is comprised of a Wikipedia
topic (title of article) and a collection of non-Wikipedia reference documents,
and the target is the Wikipedia article text.
We describe the first attempt to abstractively
generate the first section, or \textit{lead}, of Wikipedia articles
conditioned on reference text. In addition to
running strong baseline models on the task, we modify the Transformer architecture 
\citep{vaswani2017attention} to only consist of a decoder,
which performs better in the case of
longer input sequences compared to recurrent neural network (RNN) and Transformer encoder-decoder models.
Finally we show our modeling improvements allow us to generate entire Wikipedia articles.

\section{Related Work}
\label{sec:related}
\subsection{Other datasets used in neural abstractive summarization}
Neural abstractive summarization was pioneered in
\citet{Rush2015}, where they train headline generation models using the
English Gigaword corpus
\citep{graff2003english}, consisting of news articles from number of publishers.
However, the task is more akin to sentence paraphrasing
 than summarization
as only the first sentence of an article is used to predict the headline,
another sentence.
RNN-based encoder-decoder models with attention (seq2seq) perform very well on this task
in both
ROUGE \citep{lin2004rouge}, an automatic metric often used in summarization,
and human evaluation
\citep{chopra2016abstractive}.

In \citet{nallapati2016abstractive}, an abstractive
summarization dataset is
proposed by modifying a question-answering dataset of news articles paired
with story highlights from Daily Mail and CNN. This task is more difficult than
 headline-generation because the information used in the highlights may
 come from many parts of the article and not only the first sentence. One downside
 of the dataset is that it has an order-of-magnitude fewer parallel examples (310k vs. 3.8M) to learn from.
 Standard seq2seq models with attention do less well, and a number
 of techniques are used to augment performance.
 Another downside is that it is unclear what the guidelines are for
 creating story highlights and it is obvious that there are significant
 stylistic differences between the two news publishers.

 In our work we also train neural abstractive models, but
 in the multi-document regime with Wikipedia.
 As can be seen in Table~\ref{dataset-sizes}, the input and output text are
 generally much larger, with significant variance depending on the article.
 The summaries (Wikipedia lead) are multiple sentences and sometimes multiple
 paragraphs, written in a fairly uniform style as encouraged
 by the Wikipedia Manual of Style\footnote{\url{https://en.wikipedia.org/wiki/Wikipedia:Manual_of_Style}}. However, the input documents may consist
 of documents of arbitrary style originating from arbitrary sources.

 We also show in Table~\ref{dataset-sizes} the ROUGE-1 recall scores of the output
 given the input, which is the proportion of unigrams/words
 in the output co-occuring
 in the input. A higher score corresponds to a dataset more amenable to
 extractive summarization. In particular, if the output is completely embedded somewhere
 in the input (e.g. a wiki-clone), the score would be 100. Given a score of only 59.2 compared
 to 76.1 and 78.7 for other summarization datasets shows that ours is the least amenable to purely extractive methods.

\subsection{Tasks involving Wikipedia}
There is a rich body of work incorporating Wikipedia for machine learning tasks, including
question-answering (\citet{hewlett2016wikireading}, \citet{rajpurkar2016squad}) and
information extraction \citep{lehmann2015dbpedia}, and text generation from 
structured data 
\citep{Lebret2016}.

The closest work to ours involving generating Wikipedia is \cite{Sauper2009},
where articles are generated extractively (instead of abstractively in our case)
from reference documents using learned
templates. The Wikipedia articles are restricted to two categories, whereas we
use all article types. The reference documents are obtained from a search engine,
with the Wikipedia topic used as query similar to our search engine references.
However we also show results with documents only found in the
References section of the Wikipedia articles.

\subsection{Transformer models} \label{related_transformer}
Previous work on neural abstractive summarization relies on RNNs as
fundamental modules, mirroring techniques successful
in machine translation (MT).
Recently, state-of-the-art MT results were obtained using a
non-recurrent architecture, called the Transformer \citep{vaswani2017attention}.
The lack of recurrence enables greater within-training-example parallelization,
at the cost of quadratic complexity in the input sequence length.
We find the Transformer transfers well to medium length, input sequence
summarization and describe modifications to better handle longer sequences.

\begin{table}[t]
\caption{Order of magnitude input/output sizes  and
	unigram recall for summarization datasets.}
\label{dataset-sizes}
\begin{center}
\begin{tabular}{lllll}
\multicolumn{1}{c}{\bf Dataset}  &\multicolumn{1}{c}{\bf Input}
	&\multicolumn{1}{c}{\bf Output}
	&\multicolumn{1}{c}{\bf \# examples}
	&\multicolumn{1}{c}{\bf ROUGE-1 R}
\\ \hline \\
	Gigaword \citep{graff2003english}         & $10^1$         & $10^1$ & $10^6$ & $78.7$\\
	CNN/DailyMail \citep{nallapati2016abstractive} & $10^2$--$10^3$          & $10^1$  & $10^5$ & $76.1$\\
	WikiSum (ours)          & $10^2$--$10^6$  & $10^1$--$10^3$ & $10^6$ & $59.2$\\
\end{tabular}
\end{center}
\end{table}

\section{English Wikipedia as a multi-document summarization dataset}
Wikipedia, being an encyclopedia, can be viewed as a collection of
summaries on various topics given by their title,
e.g. "Canada" or "Machine Learning". The source material to be 
summarized can be viewed as all reputable documents on the Web or
books; however, to make the problem more tractable we consider the following
subsets of all documents, $D$: 

\begin{enumerate}
    \item Cited sources:
A Wikipedia article that conforms to the style
guidelines should be well-supported by citations found in the
\textit{References} section of Wikipedia articles.
For each article, $a_i$, we extract all text without
markup from crawlable citation documents, $C_i \subset D$, to use as input
to our method.
\item Web Search results:
To expand the collection of reference documents, we crawl the search results
from the Google search engine, using the article section titles as queries.
For each query, we collect 10 result pages. From this collection we remove
the Wikipedia article itself, which is often among the top results.
We also remove "clones", which are detected
when there is a high-level of unigram overlap with the article (details provided in \ref{clones}).
We denote these refined search results for an article, $a_i$, as $S_i \subset D$.
Similar to $C_i$, we extract only the text to use as input.
\end{enumerate}

\begin{table}[t]
\caption{Percentiles for different aspects of WikiSum dataset. Size is in number of words.}
\begin{center}
\begin{tabular}{rrrrrrr}
\multicolumn{1}{r}{\bf Percentile}&\multicolumn{1}{r}{\bf 20}&\multicolumn{1}{r}{\bf 40}&\multicolumn{1}{r}{\bf 50}&\multicolumn{1}{r}{\bf 60}&\multicolumn{1}{r}{\bf 80}&\multicolumn{1}{r}{\bf 100}\\ \hline \\
Lead Size	&37	&62	&78	&98	&166	&10,034 \\
Num Citations	&1	&2	&2	&3	&5	&1,029 \\
Citations Size	&562	&1,467	&2,296	&3,592	&10,320	&6,159,463 \\
Num Search Results	&10	&20	&26	&31	&46	&2,095 \\
Search Results Size	&1,1691	&33,989	&49,222	&68,681	&135,533	&5,355,671 \\
\end{tabular}
\end{center}
\label{wikisum-stats}
\end{table}

Table \ref{wikisum-stats} describes overall properties of our WikiSum dataset.
Many articles have few citations, motivating our supplementation of the source
documents with web search results. On the other hand, citations when
available, tend to be of higher-quality.  When counting the total words
in the entire dataset, it is orders-of-magnitude larger than previous summarization
datasets.

To have consistent train/development/test data across corpus-comparison experiments,
 we restrict the articles to those with at least one crawlable citation.
We divide the articles roughly into 80/10/10 for train/development/test subsets,
resulting in 1865750, 233252, and 232998 examples respectively.

\section{Methods and Models}
Because the amount of text in input reference documents $(C_i, S_i)$
can be very large
(see Table \ref{wikisum-stats})
it is infeasible to train an end-to-end abstractive model given the memory constraints
of current hardware.
Hence, we first coarsely select a subset of the input using
extractive summarization.
The second stage involves training an abstractive model that generates
the Wikipedia text while conditioning on this extraction. This two-stage
process is inspired by
by how humans might summarize multiple long documents: First highlight
pertinent information, then conditionally generate the summary based on
the highlights.

\subsection{Extractive stage}
We investigate three  extractive methods from the summarization literature,
along with a trivial and cheating method,  to assess the importance of this stage.
For each article, $a_i$ we create a ranked list of paragraphs,
$\{p^i_{R_i(j)}\}$, occurring in $(C_i, S_i)$ where $R_i(j)$ is the rank of the $j$th paragraph $p^i_j$ of $(C_i, S_i)$.
From this we select the first $L$ tokens as input to the second
abstractive stage.

\begin{enumerate}
    \item \textit{Identity}: As a trivial baseline extractor, we simply use the first $L$ tokens of the input.
\item \textit{tf-idf}: A non-trivial ranking is to consider ranking paragraphs as documents in a query-retrieval
problem, where the query is the title of the article, $T(a_i)$. We compute tf-idf \citep{ramos2003using}
for the query, with respect to the documents, $\{p^i_j\}$. That is, we summate
for each word in the query
\[N_w \cdot log (\frac{N_d}{N_{dw}}) \]
where $N_w$, $N_d$, and $N_{dw}$ are the count of the word in the document,
total number of documents, and total number of documents containing the word, 
respectively.

\item \textit{TextRank} \citep{textrank}: A weighted graph is defined
	where text units are nodes and edges are defined
		by a similarity measure based on word overlap.
	An algorithm similar to PageRank \citep{pagerank} is then used to
	compute the ranking of text units. We used paragraphs for the text units.
\item \textit{SumBasic} \citep{sumbasic}: Word frequencies in the input text
	are used to assign scores to words, which are in turn used to score
        sentences. After selecting the best scoring sentence, words in it
        have their scores reduced, and the process is repeated until the desired
	summary length is reached.

\item \textit{Cheating}
    To further demonstrate the quality of extraction on the final performance,
    we implement a cheating extractor that ranks $\{p^i_j\}$ using recall of bigrams
    in the ground truth text:

\begin{equation}
d(p^i_j, a_i) = \frac{bigrams(p^i_j) \cap bigrams(a_i)}{bigrams(a_i)}
\label{cheating-similarity}
\end{equation}
\end{enumerate}

\subsection{Abstractive stage}
\subsubsection{Data representation}
Given the ordered paragraphs $\{p^i_{R_i(j)}\}$,
we derive the raw text input simply as the concatenation of the paragraphs in
order,
the most relevant at the beginning, and prefixed with the title.

We then encode the text using sub-word tokenization similar to \citet{wu2016google}
with a vocabulary size of 32,000 yielding tokenized input, $x_i$:
   \[ text_i = T(a_i) \Vert \{p^i_{R_i(j)}\}  \]
 \[ tokenize(text_i) = x_i = (x_i^1, x_i^2, ..., x_i^{n_i}) \]

For various values of $L$ in experiments, we truncate the tokens to form the input sequence:
  \[ m^L_i = (x_i^1, ... x_i^{min(L, n_i)}) \]

For the output, we use the same vocabulary and tokenization for the Wikipedia
lead text but do not do any truncation across experiments.

Next we describe the abstractive models, $W$, that learn to write articles, $a_i = W(m^L_i)$,
which we treat as a sequence transduction problem from very long input sequences
(up to $L=11000$) to medium output sequences (typically less than 500).

\subsubsection{Baseline Models}

As a baseline we apply the standard LSTM encoder-decoder with
attention (seq2seq-att) as in \cite{bahdanau2014neural} to this task.
As is typical we train to optimize the maximum-likelihood objective:

\[ y_i = tokenize(a_i) \]
\[\prod_{i=1}^{N}p(y_i | m^L_i) \]

A stronger, more recent baseline that we use is the non-recurrent
Transformer model described in \ref{related_transformer}, which also has symmetric
encoder and decoder modules (T-ED).

\subsubsection{Transformer Decoder (T-D)}
We introduce a simple but effective modification to T-ED for long sequences
that drops the encoder module (almost reducing model parameters by half for a given
hyper-parameter set), combines the input and output
sequences into a single "sentence" and is trained as a standard language model. 

That is, we convert a sequence-transduction example $(m^1, ..., m^n) \mapsto (y^1, ..., y^\eta)$ into
the sentence $(w^1, ..., w^{n+\eta+1}) = (m^1, ..., m^n, \delta, y^1, ..., y^\eta)$, where $\delta$ is a special separator token
 and train a model to predict the next word given the previous ones:
 \[p(w^1, ..., w^{n+\eta}) = \prod_{j=1}^{n+\eta} p(w^i | w^1, ..., w^{j-1}) \]

Since the model is forced to predict the next token in the input, $m$, as well as $y$,
error signals are propagated from both input and output time-steps during training. We also
suspect that for monolingual text-to-text tasks redundant information is re-learned about language in the encoder and decoder. 
 We believe this allows for easier optimization and empirically observe this 
 with longer sequences (see Section \ref{discussion}).
 Note that because of the
 self-attention of the Transformer, when generating the next token, attention
 from both $m$ and $y$ are considered.
 At inference we provide the input sequence, $m_i$, initially, and auto-regressively generate the output, $y_i$, as normal.

\subsubsection{Transformer Decoder with memory-compressed attention (T-DMCA)} \label{t-dmca}
To re-use the terminology used to describe the Transformer,
the attention is a function of a query ($Q$) and set of key ($K$) and value ($V$) pairs.
To handle longer sequences, we modify the multi-head self-attention
of the Transformer to reduce memory usage
by limiting the dot products between $Q$ and $K$ in:
	\[Attention(Q, K, V) = softmax(\frac{QK^T}{\sqrt{d_k}}) V \]

\textbf{Local attention}: Sequence tokens are divided into blocks of similar length and 
    attention is performed in each block independently. As the attention memory cost per block 
    becomes constant, this modification allow us to keep the number of activations linear
    with respect to the sequence length. In our experiments, we choose to have blocks of 256 tokens.

\textbf{Memory-compressed attention}:
After projecting the tokens into the query, key, and value embeddings, we reduce the number of keys
and values by using a strided convolution. The number of queries remains unchanged.
This modification allows us to divide the number of activations by a compression factor.
In our experiments we use convolution kernels of size 3 with stride 3.
In contrast to local attention layers,
which only capture the local information within a block,
the memory-compressed attention layers are able to exchange information
globally on the entire sequence.

\begin{figure}[h]
\fbox{%
    \includegraphics[width=\textwidth]{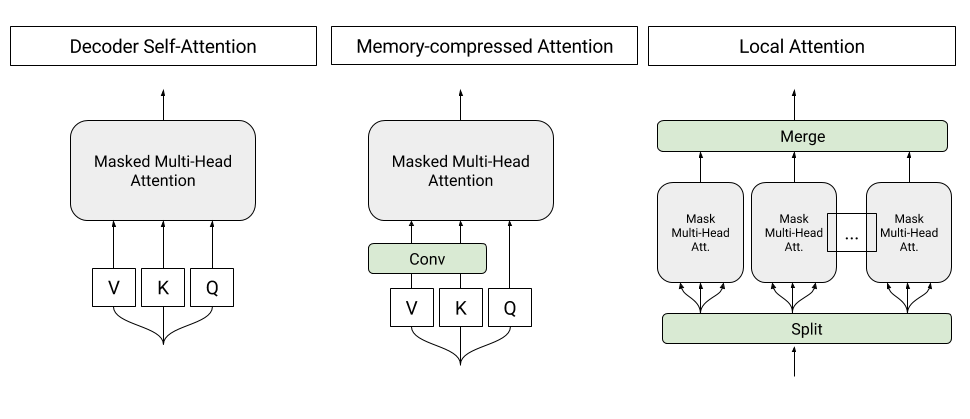}
}%
	\caption{The architecture of the self-attention layers used in the T-DMCA model.
	Every attention layer takes a sequence of tokens as input and produces
	a sequence of similar length as the output.
	\textbf{Left:} Original self-attention as used in the transformer-decoder.
	\textbf{Middle:} Memory-compressed attention which reduce the number of keys/values.
	\textbf{Right:} Local attention which splits the sequence into individual smaller sub-sequences.
	The sub-sequences are then merged together to get the final output sequence.}
\label{td_model}
\end{figure}

These modifications (see Figure \ref{td_model}) allow us in practice to process
sequences 3x in length over the T-D model.  For both local and memory-compressed attention, masking is added to prevent the queries from attending to future keys and values.
Our final architecture is a 5-layer network (LMLML) alternating between local-attention (L) layers and
memory-compressed attention (M) layers (in \citet{vaswani2017attention} it is 6 identical layers).
We also added in some experiments
one mixture of experts (MoE) layer \citep{shazeer2017outrageously}
to increase the network's capacity.

\section{Experiments}

\subsection{Evaluation}
In experiments we evaluate based on perplexity (per-wordpiece), a common language
modeling metric, and ROUGE-L F1 (version \texttt{ROUGE-1.5.5}), a common metric used in comparing
candidate and reference summaries. Note the F1 flavor of ROUGE is more 
appropriate in this setting as we do not explicitly constrain the output length in abstractive models;
it is the harmonic
mean of ROUGE-Recall (which favors long summaries) and ROUGE-Precision (which favors short
summaries).

Although optimizing ROUGE directly has been shown to not always yield the best summaries as evaluated by
human judgment \citep{paulus2017deep}, we found that for our task optimizing for perplexity
correlates with increased ROUGE and human judgment. We suspect that the relatively uniform style
of Wikipedia articles makes ROUGE more appropriate here than in general abstractive summarization tasks.

\subsection{Model Training Details and Decoding}
For all abstractive model training, we use the open-source
\texttt{tensor2tensor}\footnote{\url{https://github.com/tensorflow/tensor2tensor}}
library. 

The seq2seq baseline had a hidden size of 128 with 2 layers (we use the hyper-parameter
set defined in the library as \texttt{lstm\_attention}).

For the Transformer encoder-decoder (T-ED), we use the hyper-parameter
set  \texttt{transfomer\_base\_v1}
and train for 1 million steps.
Models exhibited very little over-fitting and did not require early-stopping.
The Transformer Decoder (T-D) was identical to the decoder part of T-ED.
The T-DMCA model is similar to T-D, but with the enhancements described in
section \ref{t-dmca}.

Unless otherwise stated, during decoding we use a beam search of size 4 and length penalty $\alpha=0.6$
\citep{wu2016google} and decode until an end-of-sequence token is reached.

\subsection{Results and Discussion}
\label{discussion}
There are four main dimensions we vary in experiments in generating Wikipedia lead sections:
\begin{enumerate}
    \item Extractive method: \textit{SumBasic}, \textit{TextRank}, \textit{tf-idf}, \textit{identity}, \textit{cheating extractor}
	\item Input corpus: \textit{citations}, \textit{search results}, \textit{combined}
	\item Abstractive model input length, $L$: We try values between 100 and 11000.
	\item Abstractive model architecture: \textit{seq2seq-att}, \textit{T-ED}, \textit{T-D}, \textit{T-DMCA}
\end{enumerate}

\begin{figure}[h]
\begin{center}
\fbox{%
    \includegraphics[width=0.6\textwidth]{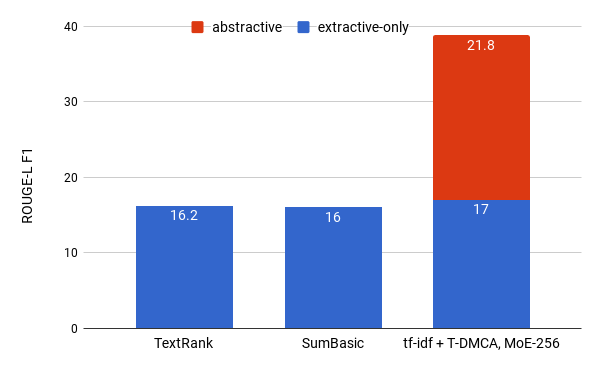}
}%
\caption{ROUGE-L F1 for various extractive methods. The abstractive model contribution is shown
for the best combined \textit{tf-idf}-T-DMCA model.}
\label{extractive_barplot}
\end{center}
\end{figure}
\textbf{Extractive-only is not enough:}
We investigate performance of extractive methods without the abstractive model by looking at the ROUGE-L F1 scores after running \textit{tf-idf}, \textit{SumBasic}, and \textit{TextRank} in
Figure \ref{extractive_barplot}, without any abstractive model.
In the case of TextRank and SumBasic we matched
the output length to the target length and observe the extractive methods perform roughly
in-line with each other in terms of ROUGE-L F1. Our best abstractive model more than doubled
this metric. Further, this model yields large improvements in perceived linguistic quality
(elaborated below).

\textbf{Extractive method:}
From Table \ref{ext-input-perf} we observe that smart extraction is critical
for final abstractive performance.
There is a significant gap between doing nothing, \textit{identity},
and extractive summarization, \textit{tf-idf}. Further, there is a
significant gap between \textit{tf-idf} and the \textit{cheating} extractor, suggesting
future work in improving the extraction step could result in significant improvements.
One possibility is to train a supervised model to predict relevance (Eq. \ref{cheating-similarity}),
which we leave as future work.
For subsequent experiments we fix the extractive method to \textit{tf-idf}.

\begin{table}[t]
\caption{Comparison of extractive method and corpus with $L=500$, and
	the Transformer E-D model}
\label{ext-input-perf}
\begin{center}
\begin{tabular}{llll}
\multicolumn{1}{l}{\bf Extractor}
	&\multicolumn{1}{l}{\bf Corpus}
	&\multicolumn{1}{l}{\bf Test log-perplexity}
	&\multicolumn{1}{l}{\bf ROUGE-L}
\\ \hline \\
	\textit{cheating} & combined & 1.72975  & 59.3 \\
	\textit{tf-idf} & combined & 2.46645 & 34.2 \\
	\textit{tf-idf} & citations-only  & 3.04299  & 22.6 \\
	\textit{tf-idf} & search-only & 3.56593  & 2.8 \\
	\textit{identity} & combined & 4.80215  & 4.0 \\
\end{tabular}
\end{center}
\end{table}

\textbf{Input Corpus:}
From table \ref{ext-input-perf} we also observe that, unsurprisingly,
the \textit{combined} dataset performs best, but the gaps between it and using
only one of \textit{citations} or \textit{search results} are both significant
and their contributions are
complementary. In subsequent experiments, we report only the \textit{combined} results.

\begin{table}[t]
\caption{Performance of best models of each model architecture
    using the combined corpus and tf-idf extractor. }
\label{abs-perf}
\begin{center}
\begin{tabular}{lll}
\multicolumn{1}{c}{\bf Model} &\multicolumn{1}{c}{\bf Test perplexity} &\multicolumn{1}{c}{\bf ROUGE-L}
\\ \hline \\
\textit{seq2seq-attention, $L=500$}          &  5.04952 & 12.7 \\
\textit{Transformer-ED, $L=500$}          & 2.46645  & 34.2 \\
\textit{Transformer-D, $L=4000$}          & 2.22216  & 33.6 \\
\textit{Transformer-DMCA, no MoE-layer, $L=11000$}          & 2.05159  & 36.2 \\
\textit{Transformer-DMCA, MoE-128, $L=11000$}          & 1.92871  & 37.9 \\
\textit{Transformer-DMCA, MoE-256, $L=7500$}          & 1.90325  & 38.8 \\
\end{tabular}
\end{center}
\end{table}
\begin{figure}[h]
\begin{center}
    \includegraphics[width=0.7\textwidth]{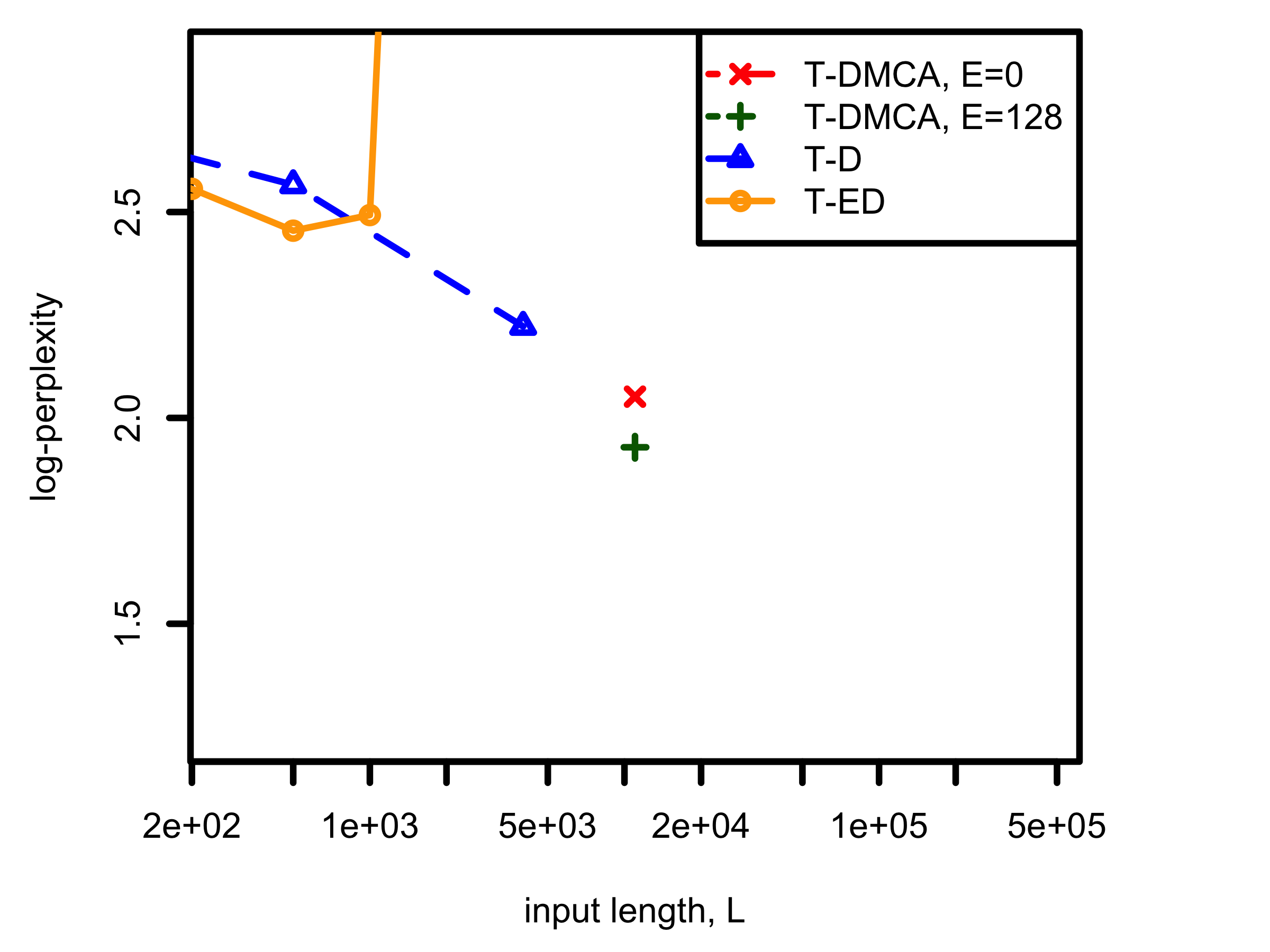}
\caption{Shows perplexity versus $L$ for tf-idf extraction on combined corpus
	for different model architectures. For T-DMCA, $E$ denotes
	the size of the mixture-of-experts layer.}
\label{perpVsL}
\end{center}
\end{figure}
\textbf{Abstractive model architecture and input length:}
As we see from Table \ref{abs-perf}, \textit{seq2seq-attention} as a baseline
does quite poorly on this task compared to the Transformer architectures.
As seen in Figure \ref{perpVsL}, we observe that the Transformer encoder-decoder, T-ED, architecture
consistently improves in performance until a best of around $L=500-1000$ and is unable to learn at $L=2000$.
This motivated the Transformer-Decoder, which we found could learn and improve up to $L=4000$, before
running out of memory on our machines equipped with 16GB of GPU RAM (NVIDIA P100).
By using the T-DMCA modifications, we were able to train up to $L=11000$ and continued to see
improvements in performance.
We also found the MoE-layer helped performance by adding model capacity at high $L$, for example
dropping log-perplexity from 2.05 to 1.93 at $L=11000$ with 128 experts. Our best model
attempted uses 256 experts at $L=7500$
(we were unable to use 256 experts with $L=11000$ due to memory constraints)
and achieves a perplexity of 1.90,

\textbf{Human Evaluation - Linguistic quality}
We conducted a DUC-style human evaluation of linguistic quality{\footnote{\url{http://duc.nist.gov/duc2007/quality-questions.txt}} of samples from 
    a baseline abstractive (seq2seq), the best extractive (\textit{tf-idf}), and our best T-DMCA models.
    Five different dimensions are assessed: grammaticality, non-redundancy, referential clarity, focus, and structure/coherence. As seen in Table \ref{ling-quality}, the T-DMCA model does statistically significantly better 
    on all dimensions, 
    except on non-redundancy where \textit{tf-idf} does about as well. Overall,
    we observed high fluency and coherence from our best abstractive model.
    Occasionally we observed some repetition of phrases which hurt
    the non-redundancy and structure, but it was much rarer compared with
    the other abstractive method, \textit{seq2seq}. The biggest weakness
    of the extractive method compared with our best abstractive model
    was the lack of structure and coherence in the summaries.

\begin{table}[t]
    \caption{Linguistic quality human evaluation scores
    (scale 1-5, higher is better). A score significantly different
(according to the Welch Two Sample t-test, with $p=0.001$)
than the \textit{T-DMCA} model is denoted by *.}
\label{ling-quality}
\begin{center}
\begin{tabular}{llllll}
\multicolumn{1}{l}{\bf Model}
    &\multicolumn{1}{l}{\bf Focus}
	&\multicolumn{1}{l}{\bf Grammar}
    &\multicolumn{1}{l}{\bf \thead{Non-\\redundancy}}
    &\multicolumn{1}{l}{\bf \thead{Referential \\ clarity}}
    &\multicolumn{1}{l}{\bf \thead{Structure and \\ Coherence}}
\\ \hline \\
\textit{T-DMCA (best)} & 4.5 & 4.6 & 4.2 & 4.5 & 4.2 \\
\textit{tf-idf}-only  & 3.0* & 3.6* & 3.9 & 3.2*  & 2.7*  \\
\textit{seq2seq-attention}  & 3.0*  & 3.4*  & 2.1*  & 3.4*  & 2.3* 
\end{tabular}
\end{center}
\end{table}

\textbf{Human Evaluation - side-by-side preference}
We validated our chosen metrics correlate with human preference by conducting
two side-by-side human evaluation experiments, comparing models with large gaps in
perplexity/ROUGE.  We observe in Table \ref{human_eval_table} that human judgment correlates with 
our automatic metrics, but it becomes more difficult to distinguish at the higher-end of model performance.
Details of the human evaluation experimental designs can be found in Appendix \ref{human_eval}.

\begin{table}[t]
\caption{Side-by-side for two models pair with large automatic metric gaps }
\label{human_eval_table}
\begin{center}
\begin{tabular}{lllll}
\multicolumn{1}{l}{\bf Model A} &\multicolumn{1}{l}{\bf Model B}
                                &\multicolumn{1}{l}{\bf ROUGE-L A}
                                &\multicolumn{1}{l}{\bf ROUGE-L B}
                                &\multicolumn{1}{l}{\bf $\frac{\textrm{\#  prefer B}}{\textrm{\# prefer A}}$}
\\ \hline \\
T-ED, $L=100$ & T-ED, $L=500$ &  30.9 & 34.2  & 4.25 \\
T-ED, $L=500$ & T-DMCA-MoE-256, $L=7500$ & 34.2 & 38.8  & 1.5 \\
\end{tabular}
\end{center}
\end{table}

To summarize the quantitative results, we believe the highest impact future work
will be from improving the extractive stage
and extending the decoder-only architectures to learn from larger $L$ while maintaining sufficient
model capacity.

\textbf{Comparison with \cite{Sauper2009}:}
A direct comparison with \cite{Sauper2009} is difficult for three reasons: (a) they report
results only for two small subsets of Wikipedia, Diseases and American Actors; 
 (b) we report on lead generation instead of full-articles;
 (c) we were unable to obtain the exact articles they used as input and output (in particular they make no claim of Wiki-clone detection). However, we
 make a best-effort comparison by finding the subset of articles
 of our test set
 that correspond to Diseases and American Actors, the two categories
 reported on by Sauper \& Barzilay and reporting our ROUGE-1 scores
 (Table \ref{sauper_table}). We observe that we perform better
 on American Actors than Diseases, probably because of the 
 prevalence of the former (and biographies) in Wikipedia compared to the latter in our training set for our single, global model, 
 whereas Sauper \& Barzilay likely benefit from the category-specific templates. On average our ROUGE-1 scores
 are higher but do worse on the less common and 
 somewhat specific disease category.
\begin{table}[t]
    \caption{Comparison of results with \cite{Sauper2009}. Note our results are reported
    for lead section, whereas Sauper \& Barzilay report for articles.}
\label{sauper_table}
\begin{center}
\begin{tabular}{llll}
\multicolumn{1}{l}{}
    &\multicolumn{1}{l}{\bf ROUGE-1 R}
    &\multicolumn{1}{l}{\bf ROUGE-1 P}
    &\multicolumn{1}{l}{\bf ROUGE-1 F1}
\\ \hline
\\
\bf All Wikipedia \\ 
\hline\\
\textit{T-DMCA} (Ours)  & 46  & 53  & 43 \\
\\
\bf Diseases \\
\hline \\
\textit{T-DMCA} (Ours), $n=161$ & 25  & 48 & 29 \\
Sauper \& Barzilay  & 36 & 39 & 37 \\
\\
\bf American Actors \\
\hline \\
\textit{T-DMCA} (Ours), $n=1322$   & 52 & 72 & 54 \\
Sauper \& Barzilay  & 46  & 40 & 41 \\ \\
\end{tabular}
\end{center}
\end{table}

\subsection{Qualitative Discussion}
\begin{figure}[h]
\begin{center}
\fbox{%
    \includegraphics[width=1.0\textwidth]{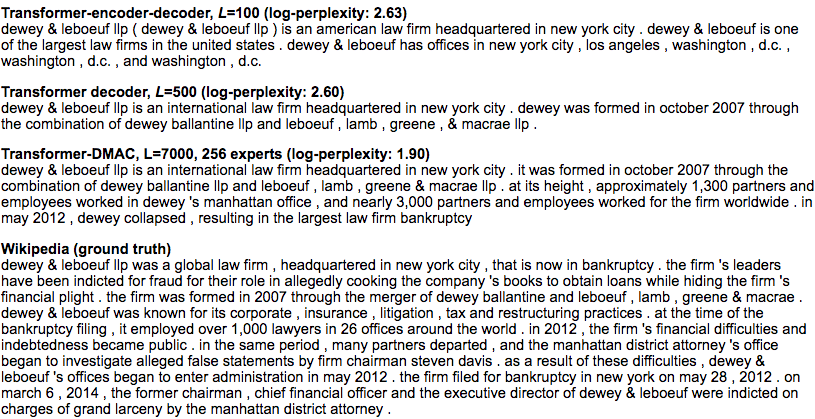}
}%
\caption{Shows predictions for the same example from different models. Example model input can
be found in the Appendix \ref{dewey_input}}
\label{compareModels}
\end{center}
\end{figure}
In Figure \ref{compareModels}, we show the predictions from three
different models (using \textit{tf-idf} extraction,
and the \textit{combined} corpus) along with the Wikipedia ground truth. As the perplexity 
decreases we see improvements in the model outputs, in terms of fluency, factual accuracy,
and narrative complexity.  
In particular, the T-DMCA model offers 
a respectable alternative to the Wikipedia version and is more succinct, while mentioning
key facts, such as where the law firm was located, when and how it was formed, and the
rise and fall of the firm.

In manual inspection of model outputs, we noticed an unexpected side-effect: models learn to
translate names from English into multiple languages, e.g. Rohit Viswanath
into Hindi (see Figure \ref{translation}). Although we did not do a systematic evaluation
of the translations, we found they are often correct, and often they are not found in the
Wikipedia article itself. We also verified that in general the translation
is not merely copied from the source, such as example cases where the target language
is the incorrect one (e.g. translation of an English name into Ukrainian).
\begin{figure}[h]
    \centering
\fbox{%
    \includegraphics[width=0.9\textwidth]{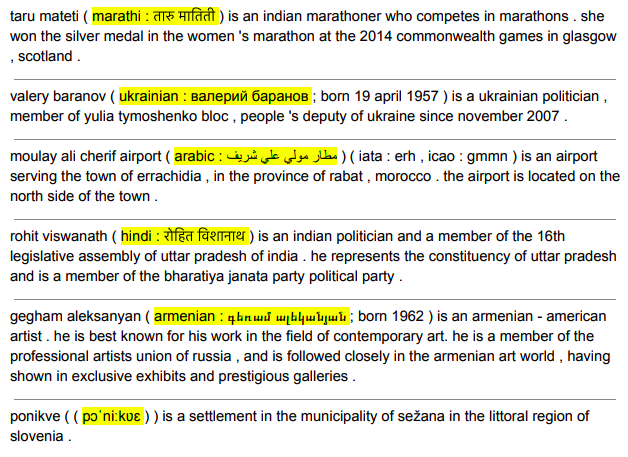}
}%
\caption{Translation examples from the Transformer-ED, $L=500$.}
\label{translation}
\end{figure}

\subsection{Generating full-Wikipedia articles}
Given that we have shown it is possible to learn sequence transduction models
on combined input-output sequence lengths of approximately 12000 using the
T-D architecture, we show that it is possible to
train a model to generate entire Wikipedia articles.
As a preliminary result, we trained two T-DMCA models: One is trained
to use $L=6000$ reference tokens to predict at most $2192$ article tokens
(longer examples are ignored) and another is conditioned only on
the title and generates articles up to $4000$ tokens long.

We show samples from both models in Appendix~\ref{full_wiki}.
Although the generated articles are not as good as the real Wikipedia
or our lead section samples,
the models can be seen to organize the article into plausible sections
and exhibit global coherence over multi-paragraph text.
The model with access to reference documents inserts factual information
in the generated article. Although we did not focus or tune on
the full-article task we see this as an interesting future work for
abstractive summarization.

\section{Conclusion}
We have shown that generating Wikipedia can be approached as a multi-document
summarization problem with a large, parallel dataset,
and demonstrated a two-stage extractive-abstractive framework for carrying it out.
The coarse extraction method used in the first
stage appears to have a significant effect on final performance, suggesting
further research on improving it would be fruitful. We introduce a new,
decoder-only sequence transduction model for the abstractive stage,
capable of handling very long input-output examples.
This model significantly outperforms traditional encoder-decoder architectures
on long sequences, allowing us to condition on many reference documents and
to generate coherent and informative Wikipedia articles.

\section{Public release of Dataset and Code}
To encourage further research on large-scale summarization,
we will release the URLs used in our experiments
(the Wikipedia URL as well as the URLs of its references)
that are available as part of the CommonCrawl dataset\footnote{\url{http://commoncrawl.org}},
which is freely available for download.

We use the open-source \texttt{tensor2tensor}\footnote{\url{https://github.com/tensorflow/tensor2tensor}}
library for
training abstractive models and will be releasing our abstractive
modeling code extensions.
Further details are available at \coderepo.

\subsubsection*{Acknowledgments}

We thank Samy Bengio, Jeff Dean, Claire Cui, Fred Bertsch, Chad Whipkey,
Anurag Rana, Ashish Vaswani, Llion Jones, and the \texttt{tensorflow/tensor2tensor}
contributors for help with the project.

\bibliography{paper_ref}

\begin{thebibliography}{2}
\providecommand{\natexlab}[1]{#1}
\providecommand{\url}[1]{\texttt{#1}}
\expandafter\ifx\csname urlstyle\endcsname\relax
  \providecommand{\doi}[1]{doi: #1}\else
  \providecommand{\doi}{doi: \begingroup \urlstyle{rm}\Url}\fi

\bibitem[Bengio \& LeCun(2007)Bengio and LeCun]{Bengio+chapter2007}
Yoshua Bengio and Yann LeCun.
\newblock Scaling learning algorithms towards {AI}.
\newblock In \emph{Large Scale Kernel Machines}. MIT Press, 2007.

\bibitem[Hinton et~al.(2006)Hinton, Osindero, and Teh]{Hinton06}
Geoffrey~E. Hinton, Simon Osindero, and Yee~Whye Teh.
\newblock A fast learning algorithm for deep belief nets.
\newblock \emph{Neural Computation}, 18:\penalty0 1527--1554, 2006.

\end{thebibliography}


\begin{thebibliography}{20}
\providecommand{\natexlab}[1]{#1}
\providecommand{\url}[1]{\texttt{#1}}
\expandafter\ifx\csname urlstyle\endcsname\relax
  \providecommand{\doi}[1]{doi: #1}\else
  \providecommand{\doi}{doi: \begingroup \urlstyle{rm}\Url}\fi

\bibitem[Bahdanau et~al.(2014)Bahdanau, Cho, and Bengio]{bahdanau2014neural}
Dzmitry Bahdanau, Kyunghyun Cho, and Yoshua Bengio.
\newblock Neural machine translation by jointly learning to align and
  translate.
\newblock \emph{arXiv preprint arXiv:1409.0473}, 2014.

\bibitem[Chopra et~al.(2016)Chopra, Auli, and Rush]{chopra2016abstractive}
Sumit Chopra, Michael Auli, and Alexander~M Rush.
\newblock Abstractive sentence summarization with attentive recurrent neural
  networks.
\newblock In \emph{Proceedings of the 2016 Conference of the North American
  Chapter of the Association for Computational Linguistics: Human Language
  Technologies}, pp.\  93--98, 2016.

\bibitem[Dang(2005)]{dang2005overview}
Hoa~Trang Dang.
\newblock Overview of duc 2005.
\newblock In \emph{Proceedings of the document understanding conference},
  volume 2005, pp.\  1--12, 2005.

\bibitem[Graff \& Cieri(2003)Graff and Cieri]{graff2003english}
David Graff and Christopher Cieri.
\newblock English gigaword 2003.
\newblock \emph{Linguistic Data Consortium, Philadeplhia}, 2003.

\bibitem[Hewlett et~al.(2016)Hewlett, Lacoste, Jones, Polosukhin, Fandrianto,
  Han, Kelcey, and Berthelot]{hewlett2016wikireading}
Daniel Hewlett, Alexandre Lacoste, Llion Jones, Illia Polosukhin, Andrew
  Fandrianto, Jay Han, Matthew Kelcey, and David Berthelot.
\newblock Wikireading: A novel large-scale language understanding task over
  wikipedia.
\newblock \emph{arXiv preprint arXiv:1608.03542}, 2016.

\bibitem[Lebret et~al.(2016)Lebret, Grangier, and Auli]{Lebret2016}
R{\'{e}}mi Lebret, David Grangier, and Michael Auli.
\newblock Neural text generation from structured data with application to the
  biography domain.
\newblock In \emph{Proceedings of the 2016 Conference on Empirical Methods in
  Natural Language Processing, {EMNLP} 2016, Austin, Texas, USA, November 1-4,
  2016}, pp.\  1203--1213, 2016.
\newblock URL \url{http://aclweb.org/anthology/D/D16/D16-1128.pdf}.

\bibitem[Lehmann et~al.(2015)Lehmann, Isele, Jakob, Jentzsch, Kontokostas,
  Mendes, Hellmann, Morsey, Van~Kleef, Auer, et~al.]{lehmann2015dbpedia}
Jens Lehmann, Robert Isele, Max Jakob, Anja Jentzsch, Dimitris Kontokostas,
  Pablo~N Mendes, Sebastian Hellmann, Mohamed Morsey, Patrick Van~Kleef,
  S{\"o}ren Auer, et~al.
\newblock Dbpedia--a large-scale, multilingual knowledge base extracted from
  wikipedia.
\newblock \emph{Semantic Web}, 6\penalty0 (2):\penalty0 167--195, 2015.

\bibitem[Lin(2004)]{lin2004rouge}
Chin-Yew Lin.
\newblock Rouge: A package for automatic evaluation of summaries.
\newblock In \emph{Text summarization branches out: Proceedings of the ACL-04
  workshop}, volume~8. Barcelona, Spain, 2004.

\bibitem[Mihalcea \& Tarau(2004)Mihalcea and Tarau]{textrank}
Rada Mihalcea and Paul Tarau.
\newblock Textrank: Bringing order into text.
\newblock In \emph{Proceedings of the 2004 conference on empirical methods in
  natural language processing}, 2004.

\bibitem[Nallapati et~al.(2016)Nallapati, Zhou, dos Santos, glar
  Gul{\c{c}}ehre, and Xiang]{nallapati2016abstractive}
Ramesh Nallapati, Bowen Zhou, Cicero dos Santos, {\c{C}}a~glar Gul{\c{c}}ehre,
  and Bing Xiang.
\newblock Abstractive text summarization using sequence-to-sequence rnns and
  beyond.
\newblock \emph{CoNLL 2016}, pp.\  280, 2016.

\bibitem[Nenkova \& Vanderwende(2005)Nenkova and Vanderwende]{sumbasic}
Ani Nenkova and Lucy Vanderwende.
\newblock The impact of frequency on summarization.
\newblock \emph{Microsoft Research, Redmond, Washington, Tech. Rep.
  MSR-TR-2005}, 101, 2005.

\bibitem[Page et~al.(1999)Page, Brin, Motwani, and Winograd]{pagerank}
Lawrence Page, Sergey Brin, Rajeev Motwani, and Terry Winograd.
\newblock The pagerank citation ranking: Bringing order to the web.
\newblock Technical report, Stanford InfoLab, 1999.

\bibitem[Paulus et~al.(2017)Paulus, Xiong, and Socher]{paulus2017deep}
Romain Paulus, Caiming Xiong, and Richard Socher.
\newblock A deep reinforced model for abstractive summarization.
\newblock \emph{arXiv preprint arXiv:1705.04304}, 2017.

\bibitem[Rajpurkar et~al.(2016)Rajpurkar, Zhang, Lopyrev, and
  Liang]{rajpurkar2016squad}
Pranav Rajpurkar, Jian Zhang, Konstantin Lopyrev, and Percy Liang.
\newblock Squad: 100,000+ questions for machine comprehension of text.
\newblock \emph{arXiv preprint arXiv:1606.05250}, 2016.

\bibitem[Ramos et~al.(2003)]{ramos2003using}
Juan Ramos et~al.
\newblock Using tf-idf to determine word relevance in document queries.
\newblock In \emph{Proceedings of the first instructional conference on machine
  learning}, volume 242, pp.\  133--142, 2003.

\bibitem[Rush et~al.(2015)Rush, Chopra, and Weston]{Rush2015}
Alexander~M. Rush, Sumit Chopra, and Jason Weston.
\newblock A neural attention model for abstractive sentence summarization.
\newblock In \emph{Proceedings of the 2015 Conference on Empirical Methods in
  Natural Language Processing, {EMNLP} 2015, Lisbon, Portugal, September 17-21,
  2015}, pp.\  379--389, 2015.
\newblock URL \url{http://aclweb.org/anthology/D/D15/D15-1044.pdf}.

\bibitem[Sauper \& Barzilay(2009)Sauper and Barzilay]{Sauper2009}
Christina Sauper and Regina Barzilay.
\newblock Automatically generating wikipedia articles: A structure-aware
  approach.
\newblock In \emph{Proceedings of the Joint Conference of the 47th Annual
  Meeting of the ACL and the 4th International Joint Conference on Natural
  Language Processing of the AFNLP: Volume 1 - Volume 1}, ACL '09, pp.\
  208--216, Stroudsburg, PA, USA, 2009. Association for Computational
  Linguistics.
\newblock ISBN 978-1-932432-45-9.
\newblock URL \url{http://dl.acm.org/citation.cfm?id=1687878.1687909}.

\bibitem[Shazeer et~al.(2017)Shazeer, Mirhoseini, Maziarz, Davis, Le, Hinton,
  and Dean]{shazeer2017outrageously}
Noam Shazeer, Azalia Mirhoseini, Krzysztof Maziarz, Andy Davis, Quoc Le,
  Geoffrey Hinton, and Jeff Dean.
\newblock Outrageously large neural networks: The sparsely-gated
  mixture-of-experts layer.
\newblock \emph{arXiv preprint arXiv:1701.06538}, 2017.

\bibitem[Vaswani et~al.(2017)Vaswani, Shazeer, Parmar, Uszkoreit, Jones, Gomez,
  Kaiser, and Polosukhin]{vaswani2017attention}
Ashish Vaswani, Noam Shazeer, Niki Parmar, Jakob Uszkoreit, Llion Jones,
  Aidan~N Gomez, Lukasz Kaiser, and Illia Polosukhin.
\newblock Attention is all you need.
\newblock \emph{arXiv preprint arXiv:1706.03762}, 2017.

\bibitem[Wu et~al.(2016)Wu, Schuster, Chen, Le, Norouzi, Macherey, Krikun, Cao,
  Gao, Macherey, et~al.]{wu2016google}
Yonghui Wu, Mike Schuster, Zhifeng Chen, Quoc~V Le, Mohammad Norouzi, Wolfgang
  Macherey, Maxim Krikun, Yuan Cao, Qin Gao, Klaus Macherey, et~al.
\newblock Google's neural machine translation system: Bridging the gap between
  human and machine translation.
\newblock \emph{arXiv preprint arXiv:1609.08144}, 2016.

\end{thebibliography}
\bibliographystyle{iclr2018_conference}

\clearpage 
\appendix
\section{Appendix}
\subsection{Examples of full wikipedia generated samples}
\begin{figure}[h]
\begin{center}
\fbox{%
    \includegraphics[width=1.0\textwidth]{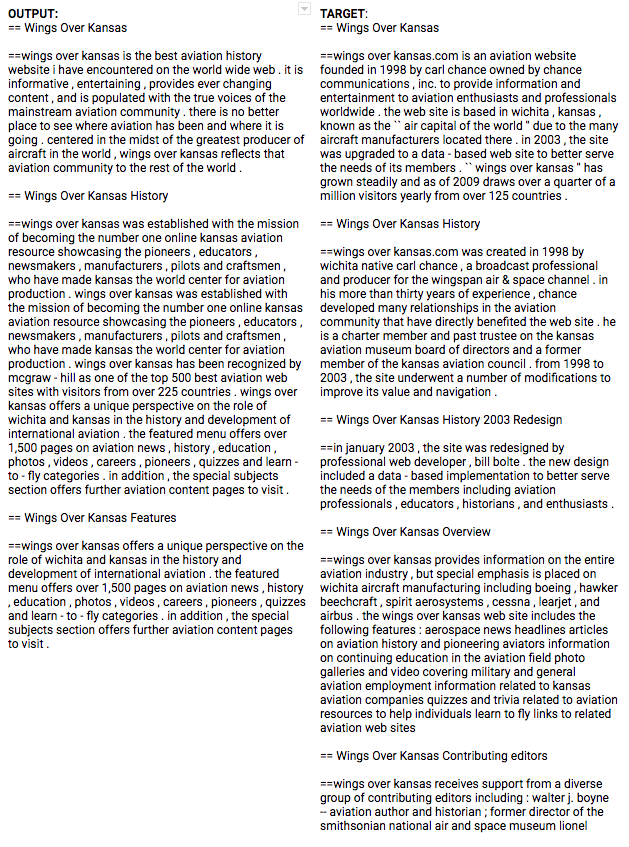}
}%
\caption{An example decoded from a T-DMCA model trained to produce an entire Wikipedia article, conditioned on 8192 reference document tokens.}
\end{center}
\label{full_wiki}
\end{figure}

\begin{figure}[h]
\begin{center}
\fbox{%
    \includegraphics[width=1.0\textwidth]{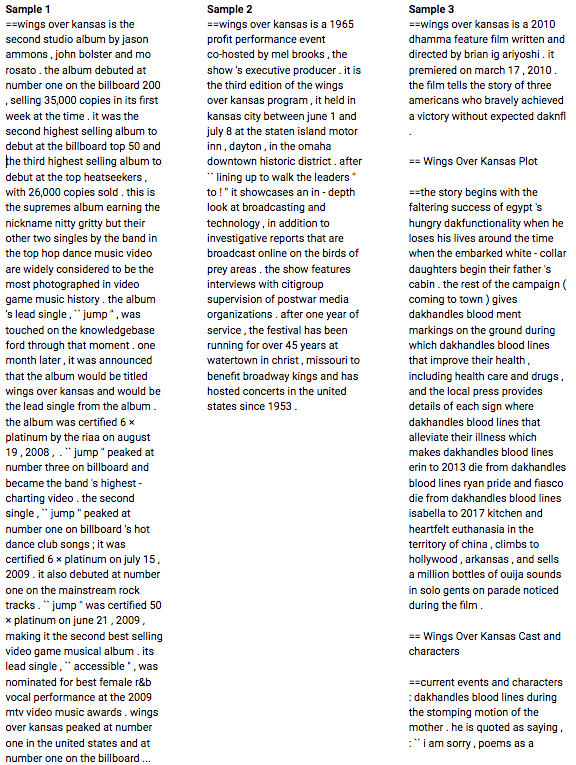}
}%
\caption{Three different samples a T-DMCA model trained to produce an entire Wikipedia article, conditioned only on the title. Samples 1 and 3 are truncated due to space constraints.}
\end{center}
\label{full_wiki_title_only}
\end{figure}

\subsection{Implementation Details} \label{implementation}
\subsubsection{Wikipedia clone detection} \label{clones}
For detecting whether a reference document, $d$, is a Wikipedia article clone
we compute the
maximum recall of unigrams for each section of the Wikipedia article, $a$:

\[
r(d, a) = \max_{s \in sections(a)} \frac{|unigrams(d) \cap unigrams(s)|}{|unigrams(s)|} \]

and detect a clone if $r>0.5$.

\subsection{Human Evaluation experiment} \label{human_eval}
\begin{figure}[h]
\begin{center}
\fbox{%
    \includegraphics[width=0.7\textwidth]{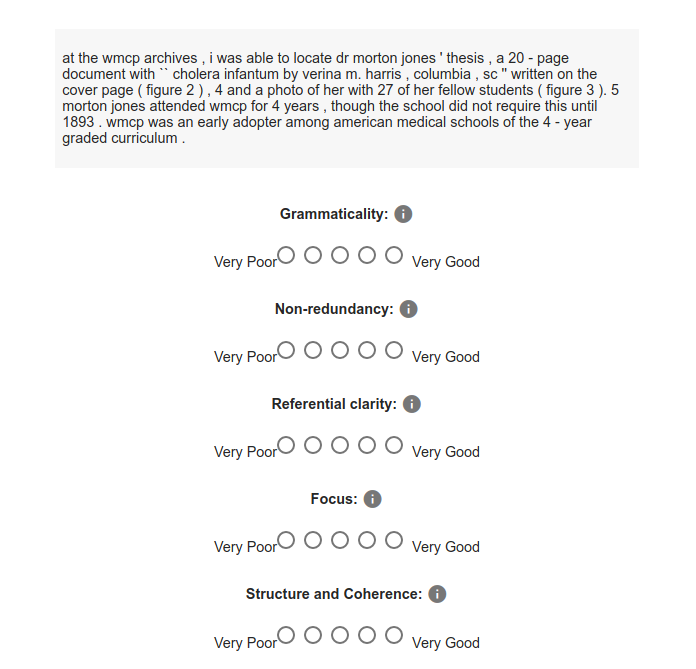}
}%
\caption{Screenshot of DUC-style linguistic quality human evaluation tool.}
\label{crowdcompute_ling}
\end{center}
\end{figure}
To assess linguistic quality, we randomly selected samples
generated by models from the test set and ask raters to choose a score 
from 1 to 5 (higher is better) for five dimensions: Grammaticality,
Non-redundancy, Referential clarity, Focus, and Structure and Coherence. These
were used in the past at DUC for evaluating summaries \citep{dang2005overview}. For each model we selected 25 examples and averaged the scores for each
question across 3 raters (out of pool of 7).

\begin{figure}[h]
\begin{center}
\fbox{%
    \includegraphics[width=0.7\textwidth]{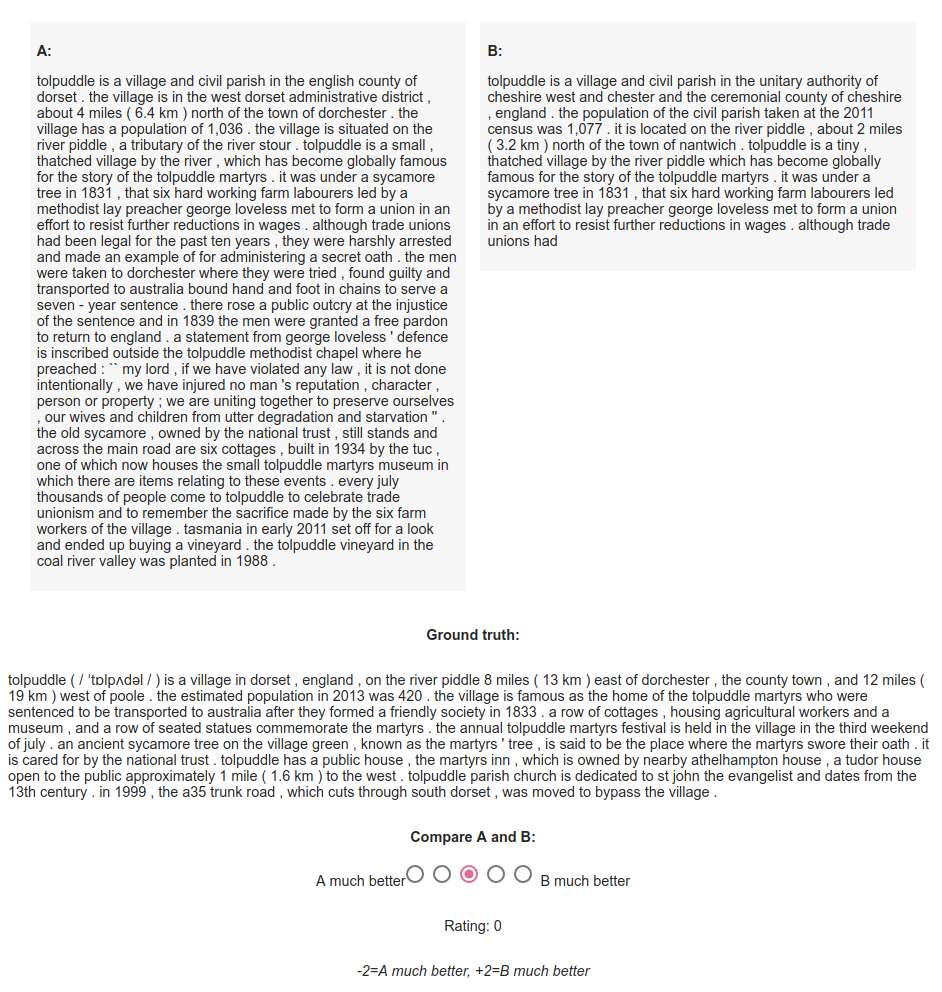}
}%
\caption{Screenshot of side-by-side human evaluation tool. Raters are asked whether they prefer
model output on the left or right, given a ground truth Wikipedia text.}
\label{crowdcompute}
\end{center}
\end{figure}
To compare two models by human evaluation, we randomly select examples from the test set and
show model outputs side-by-side in the interface shown in Figure \ref{crowdcompute}. 
Which side a model appears on is randomized per example and rater. For the
experiments in Table \ref{human_eval_table} we had 3 raters score 25 examples each
and computed the ratio of ratings preferring one model over the other.

\subsection{Example abstractive model input}
\begin{figure}[h]
\begin{center}
\fbox{%
    \includegraphics[width=1.0\textwidth]{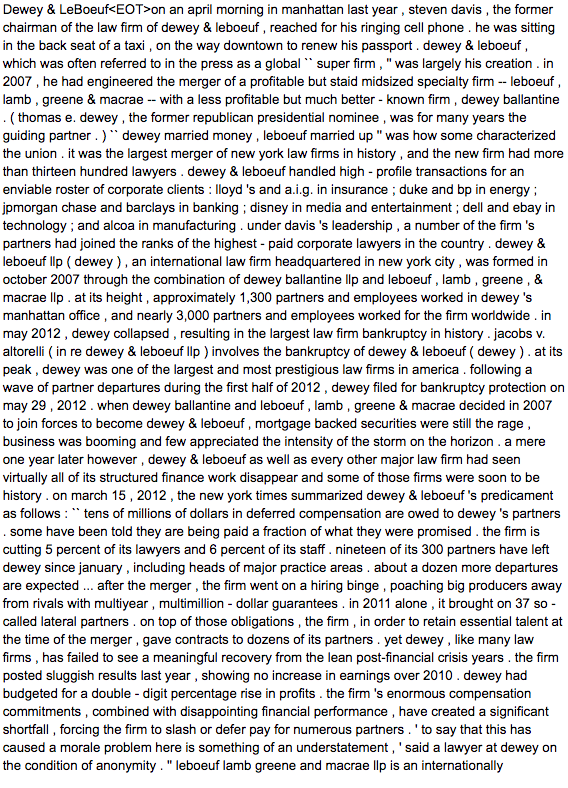}
}%
\caption{Example extractive-output/abstractive-input for models in "dewey \& lebeouf" example. The
	extractive method used is \textit{tf-idf}.}
\end{center}
\label{dewey_input}
\end{figure}

\end{document}